\documentclass[letterpaper, 10 pt, conference]{ieeeconf}  %

\IEEEoverridecommandlockouts                              %

\usepackage{graphicx}
\usepackage{bm}
\usepackage{amsmath} %
\usepackage{amssymb}  %
\usepackage{multirow}
\usepackage{todonotes}
\setuptodonotes{inline}
\usepackage{url}
\usepackage{algorithm}
\usepackage{algpseudocode}
\usepackage{siunitx}
\usepackage[hidelinks]{hyperref}
\usepackage{cleveref}
\usepackage{booktabs}

\title{\LARGE \bf
Getting the Ball Rolling: 
Learning a Dexterous Policy\\ for a Biomimetic Tendon-Driven Hand with Rolling Contact Joints
}

\author{Yasunori Toshimitsu$^{1,2}$,  %
        Benedek Forrai$^{1}$,  %
        Barnabas Gavin Cangan$^{1}$,  %
        Ulrich Steger$^{1}$,  %
        Manuel Knecht$^{1}$,  %
        \\
        Stefan Weirich$^{1}$,  %
        Robert K. Katzschmann$^{1}$  %
\thanks{$^{1}$ETH Zurich
        {\tt\footnotesize \{\href{mailto:ytoshimitsu@ethz.ch}{ytoshimitsu},
        \href{mailto:bforrai@ethz.ch}{bforrai},
        \href{mailto:bcangan@ethz.ch}{bcangan},
        \href{mailto:ulsteger@ethz.ch}{ulsteger},
        \href{mailto:knechtma@ethz.ch}{knechtma},
        \href{mailto:sweirich@ethz.ch}{sweirich},
        \href{mailto:rkk@ethz.ch}{rkk}\}@ethz.ch}}%
\thanks{$^{2}$Max Planck ETH Center for Learning Systems}%
}

\begin{document}

\maketitle
\thispagestyle{empty}
\pagestyle{empty}

\begin{abstract}
Biomimetic, dexterous robotic hands have the potential to replicate much of the tasks that a human can do, and to achieve status as a general manipulation platform. Recent advances in reinforcement learning (RL) frameworks have achieved remarkable performance in quadrupedal locomotion and dexterous manipulation tasks. Combined with GPU-based highly parallelized simulations capable of simulating thousands of robots in parallel, RL-based controllers have become more scalable and approachable.
However, in order to bring RL-trained policies to the real world, we require training frameworks that output policies that can work with physical actuators and sensors as well as a hardware platform that can be manufactured with accessible materials yet is robust enough to run interactive policies.
This work introduces the biomimetic tendon-driven \textit{Faive Hand} and its system architecture, which uses tendon-driven rolling contact joints to achieve a 3D printable, robust high-DoF hand design.
We model each element of the hand and integrate it into a GPU simulation environment to train a policy with RL, and achieve zero-shot transfer of a dexterous in-hand sphere rotation skill to the physical robot hand.\footnote{\url{https://srl-ethz.github.io/get-ball-rolling/} video: \url{https://youtu.be/YahsMhqNU8o}}
\end{abstract}

\section{Introduction}

\subsection{Motivation}
As robotic structures get more complex and biomimetic, we are starting to apply policies trained through reinforcement learning instead of conventional model-based control methods in which the controller explicitly reasons with the dynamic model of the robot. This is especially the case for dexterous manipulation tasks, which apply an anthropomorphic robotic hand to movements that require the coordination of multiple fingers. Achieving this coordinated
motion has the potential to replace many repetitive tasks such as pick-and-place in warehouses, assembly in factory lines, or assistance in our daily lives.

In this work, we introduce the Faive Hand, a platform for dexterous manipulation tasks. We report our current progress on integrating its model into an RL environment, and apply a closed-loop controller on the real robot to achieve dexterous in-hand sphere rotation as first step towards humanlike manipulation.

\begin{figure}[ht]
        \centering
        \includegraphics[width=\columnwidth]{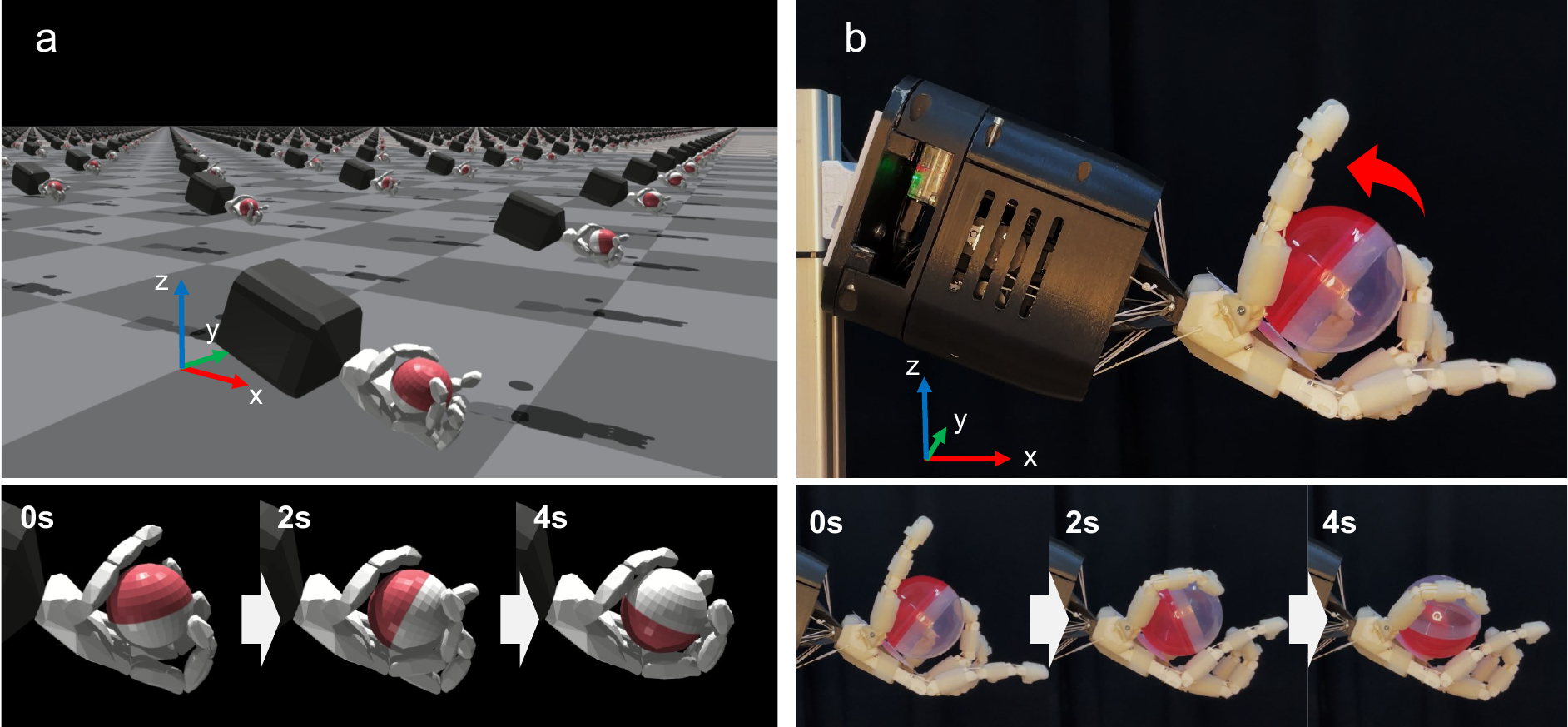}
        \caption{
        \textbf{(a)} The GPU-based parallelized simulation environment simulating 4096 robot hands in parallel to train a RL policy.
        \textbf{(b)} The trained policy being deployed on the tendon-driven robot hand with rolling contact joints.
        }
        \label{fig:humanoids_overview}
\end{figure}

\subsection{State of the art in dexterous manipulation}
\paragraph{Learning-based control}
Dexterous manipulation tasks benefit from learning-based control approaches as model-based approaches struggle with the number of different contact states that the controller needs to handle. As contact can occur anywhere along the link chain, the number of contact states can be several orders of magnitudes greater than the number of states for other tasks such as locomotion. The first major breakthrough in applying learning-based methods to dexterous manipulation on real robots was achieved by OpenAI for in-hand cube rotation on the Shadow Hand \cite{OpenAI2018-te}. However, the combination of a CPU-based simulator with a sample-inefficient RL algorithm required up to 50 hours of computation on 384 machines with 16 CPU cores each to run in parallel to collect on-policy experience, which limited the scalability to different tasks and the accessibility of replicating the same setup at other institutions. Since then, the introduction of GPU-based simulators like IsaacGym \cite{Makoviychuk2021-ll}, which can simulate thousands of robots in parallel, have greatly brought down the required resource for training RL agents. Some tasks such as locomotion can even be trained on the order of minutes on a single GPU \cite{Rudin2022-xe}. Though the tasks are not directly comparable, all of the dexterous manipulation policies in this paper were also learned in about one hour on a single NVIDIA A10G GPU. 

There has been much recent progress on applying manipulation policies learned in a parallel simulation environment such as IsaacGym onto physical robotic hands. Handa et al. proposed a vision-based policy together with a vectorized implementation of automatic domain randomization (ADR) which enabled the cube rotation task to be run on the Allegro hand using only RGB cameras for exteroceptive input \cite{Handa2022-sn}. Chen et al. created a policy that can reorient multiple objects in simulation \cite{Chen2022-ag}. Yin et al. then proposed a pipeline to use just proprioceptive touch sensor inputs to rotate objects in-hand around a desired axes \cite{Yin2023-kv}. Allshire et al. used IsaacGym to achieve dexterous manipulation on the TriFinger robot system \cite{Allshire2022-ik}.

\paragraph{Robotic hand hardware}
It is important to note that capable robots require a combination of hardware and controller, and we will take a look at the major robotic hands being used for dexterous manipulation research.
Most of the previous works use either the Allegro hand \cite{WonikRobotics2023}, a larger-than-life four-finger hand that contains servo motors in each joint, or the Shadow hand \cite{ShadowRobot2023}, which is an anthropomorphic five-finger tendon-driven robotic hand. Further, there have also been other robotic hands created at research institutions. The LEAP Hand has a similar design to the Allegro hand with an improved joint layout and robustness \cite{Shaw_undated-sg}. The TriFinger robot used in IsaacGym-based RL tasks \cite{Allshire2022-ik} and offline RL competitions \cite{Gurtler2023-hv} is a non-anthropomorphic three-finger manipulator which uses BLDC motors \cite{Wuthrich2021-cb}. 

We argue that for achieving manipulation in human environments, it is more advantageous to be closer to the human form: tools and objects in our environment are originally designed to be used by humans, so a human-like hand design is more suited for interacting with them. Also, when learning from human demonstrations, manipulation tasks can be more easily transferred to a robot with a similar structure. Robotic hands with servo motors driving each axis, like the Allegro hand, are simple to construct (consequently lowering the cost), but their fingers become substantially thicker than human ones. The Shadow Hand is by far the most human-like hand, but it comes with a steep price tag of 110k GBP as quoted from their website, limiting accessibility to conduct physical experiments. The high purchase and maintenance costs may discourage frequent sim2real experiments of RL trained policies, since they can initially behave erratically on the real robot and require intense trial-and-error until they can work fully in reality.

\subsection{Approach}
At the Soft Robotics Lab, we have developed the \textit{Faive Hand}, a biomimetic dexterous tendon-driven robotic platform for exploring dexterous manipulation. The current version of the hand uses 3D printed components and servo motors for accessible and simple manufacturing.
However, in addition to the challenges inherent to controlling a high-DoF robotic hand for manipulation, this hand has features that do not exist in other dexterous hands trained with RL, such as rolling contact joints that rotate without a fixed axis of rotation. Conventional rotational encoders are difficult to use on this design, so the hand currently does not have internal joint angle encoders, which are being developed for a later version of the hand. Because of this limitation, the joint angles must be estimated from the tendon length, which can be calculated from the servo motor angles. These features were implemented in the simulation framework and on the low-level controller, which enabled running a closed-loop RL trained policy on the physical robot. We have chosen a task similar to Shi et al. \cite{Shi2021-gy}, where the robot dexterously rotates a sphere in the target direction.

\subsection{Contributions}
\begin{itemize}
    \item Integrate a model for rolling contact joints into the IsaacGym simulator;
    \item Apply a closed-loop policy trained in simulation to control the tendon-driven biomimetic Faive Hand; and
    \item Introduce the prototype version of the \textit{Faive Hand}, designed as an accessible platform for dexterous manipulation.
\end{itemize}

\section{The tendon-driven dexterous hand with robust rolling contact joints}

In this section, we introduce the biomimetic joint structure of the \textit{Faive Hand} \footnote{\url{https://www.faive-robotics.com/} for inquiries on using the hand in your projects, please use the contact form in the website.} and how it was modelled and controlled so that RL policies can be run on the real robot. The Faive Hand was developed in our lab to research biomimetic manipulation, with the aim to eventually provide a low-cost platform that makes dexterous manipulation research on real hardware accessible to many research institutes, accelerating the application of anthropomorphic robotic hands to real-life applications.

\subsection{Rolling contact joint design of the hand}
\begin{figure}
        \centering
        \includegraphics[width=0.9\linewidth]{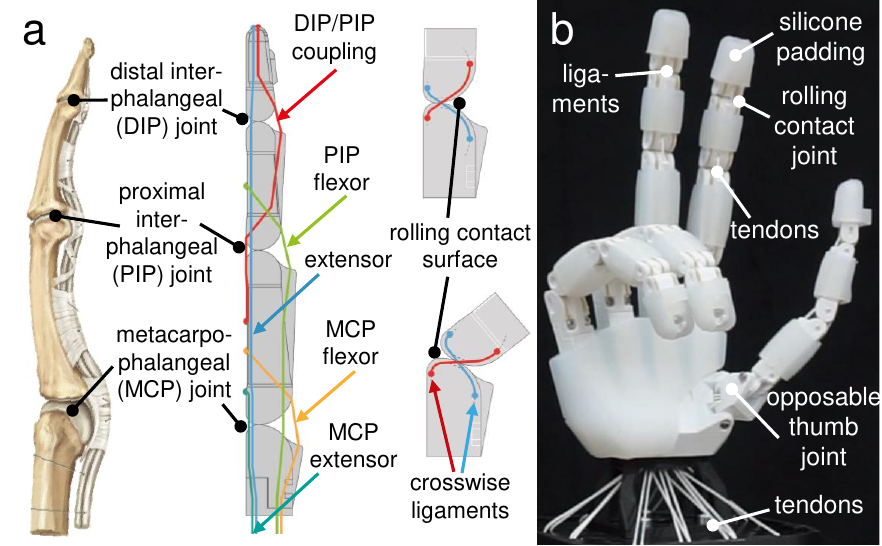}
        \caption{
        \textbf{(a)} The Faive Hand has a rolling contact joint design with tendons and ligaments mimicking that of a human finger.
        \textbf{(b)} Overview of each component of the Faive Hand.}
        \label{fig:faive_hand_overview}
\end{figure}

Here, we describe the prototype version of the \textit{Faive Hand} used for experiments in this paper, with the designation \textit{Proto 0}. It contains 11 actuatable degrees of freedom, with 3 in the thumb and 2 for each of the other fingers. Each finger contains a coupled joint at the distal end, and thus there are 16 joints in total. Similar to a human finger, our robotic finger design consists of three joints for which we apply a joint naming convention derived from human anatomy, as shown in \cref{fig:faive_hand_overview}. For each finger, the distal interphalangeal (DIP) joint  is linked to the proximal interphalangeal (PIP) joint using a coupling tendon so that they bend together. Therefore, the DIP and PIP joint are driven by one flexor tendon and a separate extensor tendon, that drive both joints simultanously.
The metacarpophalangeal (MCP) joint is actuated antagonistically by a single motor, to which both the flexor and extensor tendons are attached.

There have been many joint designs that have been used for biomimetic articulated hands, such as pin joints \cite{ShadowRobot2023, Salisbury1982-xs}, machined springs \cite{Kawaharazuka2020-ff, Makino2018-ki}, soft mechanisms \cite{Puhlmann2022-yi, Schlagenhauf2018-xt}, and rolling contact joints \cite{Kim2023-cx, Kim2019-vc}.

Apart from the carpometacarpal joint of the thumb which is recreated using 2 hinge joints, all of the joints are implemented as rolling contact joints. These rolling joints are composed of two articulating bodies with adjacent curved contact surfaces connected by a pair of crosswise ligament strings as shown in \cref{fig:faive_hand_overview} (like the children's toy Jacob's ladder). They have advantages such as impact compliance, low friction or greater range of motion, and have been proposed for robotics, implants, and prosthetics \cite{noauthor_undated-ir, Kim2016-fl}. There have also been extensions to rolling contact joints proposed with fluid lubricated joints encased in artificial skin \cite{Kim2019-vc}, or with variable stiffness \cite{Kim2023-cx}.

\begin{figure}
        \centering
        \includegraphics[width=0.9\linewidth]{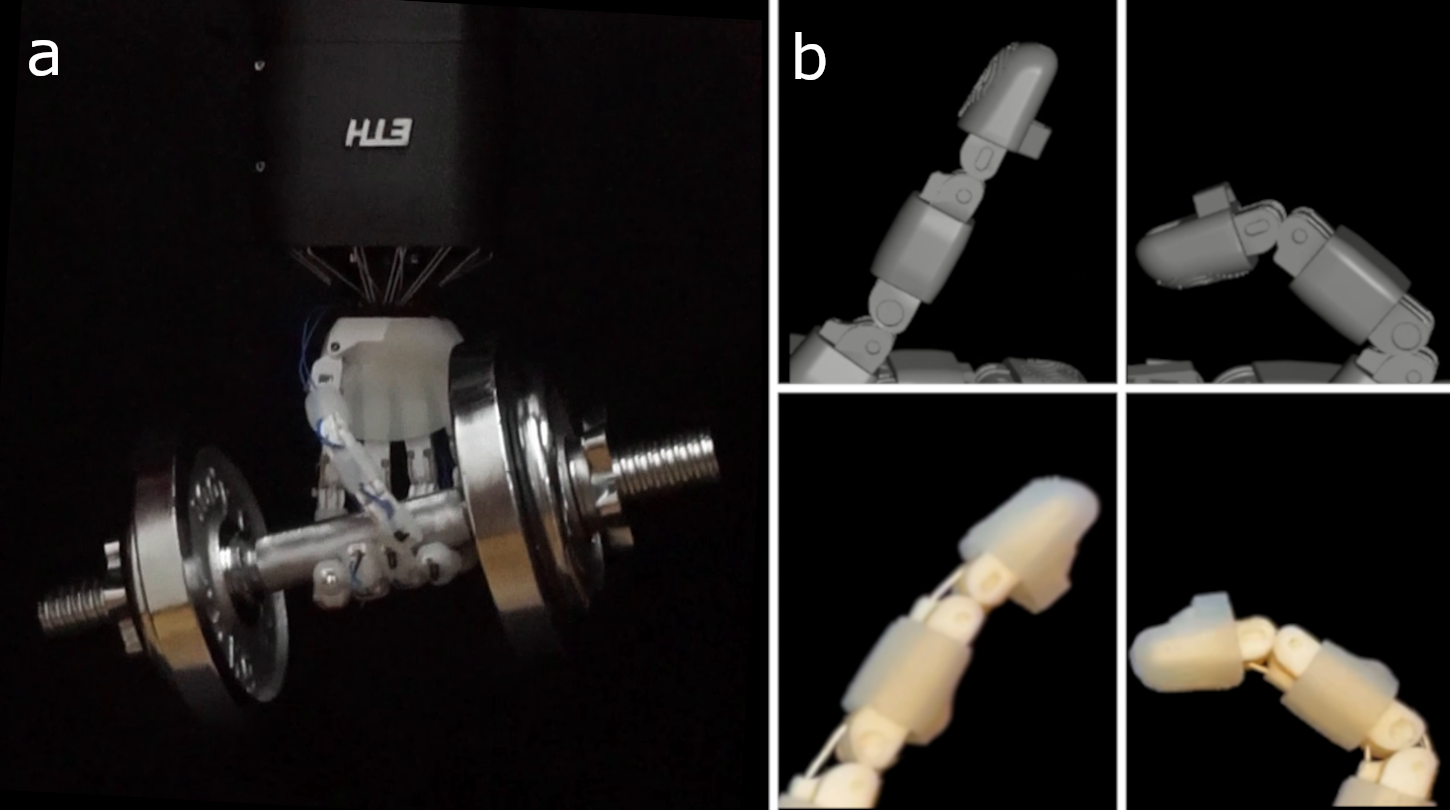}
        \caption{
        \textbf{(a)} The Faive Hand can grasp a payload of up to \SI{10}{\kilogram}, demonstrated here with a dumbbell.
        \textbf{(b)} The motion of the rolling contact joint in the MuJoCo simulator and on the real robot hand, which do not rotate around a fixed axis.}
        \label{fig:payload_rolling_contact_sim}
\end{figure}

Our finger design builds on these previous works while further simplifying the design, making it more robust, compact and easier to manufacture. We demonstrated payloads of up to \SI{10}{\kilogram} using a downwards facing five-fingered power grasp, at a total weight of the hand system of only \SI{1.1}{\kilogram}, as shown in \cref{fig:payload_rolling_contact_sim}.

\subsection{Modelling rolling contact joints in simulation}
As seen in \cref{fig:payload_rolling_contact_sim}, these rolling contact joints do not have a single axis of rotation. Thus they were modelled in simulation with two "virtual" hinge joints. The axes of these hinge joints were placed to go through the axis of the cylinder that constitutes each rolling contact surface.
They were constrained to rotate together in the MJCF (the modeling format for MuJoCo) file by linking them together with \texttt{tendon/fixed} elements, which apply constraints to the linear combination of joint angles. The tendon model information is carried over when the MJCF file is loaded in IsaacGym, and can be enabled by setting the \texttt{limit\_stiffness} and \texttt{damping} values to each of the tendon properties. 

\subsection{Low-level controller to enable joint control and sensing}
\label{sec:low_level_controller}
Similar to how the Shadow Hand was modelled in OpenAI's work \cite{OpenAI2018-te}, the hand was simulated as a joint-driven robot, ignoring the tendons-level information. Instead, the low-level controller of the robot ran a program that converts joint commands and measurements to and from their tendon counterparts.

\paragraph{Converting joint angle commands to motor-level commands}
The tendons are controlled by 16 \textit{Dynamixel XC330-T288-T} servo motors. 6 of them have two tendons attached antagonistically. Having more than one tendon on one motor will reduce the number of servos needed, but can only be used when the motion of the two tendons always have a constantly scaled relation to each other, expressed by the ratio of their spool radius to each other. This is true for the proximal joints of the finger\footnote{In theory, due to a varying moment arm, the anatagonistic relation slightly varies, but in practice this variation is small enough to be negligible.}. However, due to the routing of the tendons, the distal joints' tendon lengths depend on the proximal joints' angle and they do not have this constant relationship. Thus, the distal joints require a dedicated motor for a single tendon.

By geometrically modeling the rolling motion of the joints and the tendon path based on CAD data, we can analytically describe the function $\bm l = f(\bm q)$ mapping the joint angles $\bm q$ to the tendon lengths $\bm l$. With this function, the desired joint angle $\bm{\bar q}$ can be converted to the desired tendon length $\bm{\bar l} = f(\bm{\bar q})$ and subsequently to the desired servo motor angles (by dividing it with the tendon spool radius), and can be sent to the Dynamixel motors.

\paragraph{Joint angle sensing with an extended Kalman filter}
The function $f(\cdot)$ maps the set of all joint angles to a manifold in the space expressing the tendon length. This is not a bijection, and this mapping cannot be easily inverted: there are combinations of tendon lengths that do not map back to a configuration in joint space.
Therefore, we adopt the method used by Ookubo et al., which uses an extended Kalman filter (EKF) for estimating the joint angles from the tendon length measurements \cite{Ookubo2015-bc}. Here,  we denote the state and observation as $\bm x$ and $\bm z$ respectively, formulated as a concatenation of the position and velocity as follows.
\begin{equation}
\bm x := 
\begin{bmatrix}
    \bm q \\ \bm{\dot q}  
\end{bmatrix}
\in \mathbb R^{22}
\quad
\bm z := 
\begin{bmatrix}
    \bm l \\ \dot{\bm l}
\end{bmatrix}
\in \mathbb R^{32}
\end{equation}
The transition model can be described as
\begin{equation}
\bm x_{i+1}
=
\begin{bmatrix}
I && I dt \\
O && I
\end{bmatrix}
\bm x_i
+ \bm w
\end{equation}
where the subscript $i$ denotes the time step, $I$ is the unit matrix, $O$ is the zero matrix, $dt$ is the time step, and $\bm w$ is the noise in the state transition.

By using a symbolic computation library such as \textit{sympy} \cite{10.7717/peerj-cs.103}, the partial derivative of the function $f(\cdot)$ mapping joint angle to tendon length can be symbolically derived, giving us the muscle Jacobian $J_m = \partial f( \bm q) / \partial \bm q$. Then, the nonlinear observation model used for the EKF can be described as
\begin{equation}
\begin{split}
    \bm z_i
    &=
    \bm h(\bm x_i)
    + \bm v
    \\
    \bm h(\bm x_i) &:= \begin{bmatrix}
        f(\bm q_i)
        \\
        J_m(\bm q_i) \dot{\bm q}_i
    \end{bmatrix}
\end{split}
\end{equation}  

A calibration procedure to set the relation between motor position (spool angle) and tendon length is run every time the robot boots. We fit a jig that constrains the robot hand joints to a known pose and lightly pull on the tendons until they are taut. The program uses the motor position at that moment as the basis from which the tendon lengths are calculated.
The estimated joint angle from the EKF $\bm{\hat q}$ can then be used as the proprioceptive measurement to the RL policy as described in the subsequent section.

\section{Reinforcement learning training for dexterous manipulation}
\begin{figure}
        \centering
        \includegraphics[width=0.9\columnwidth]{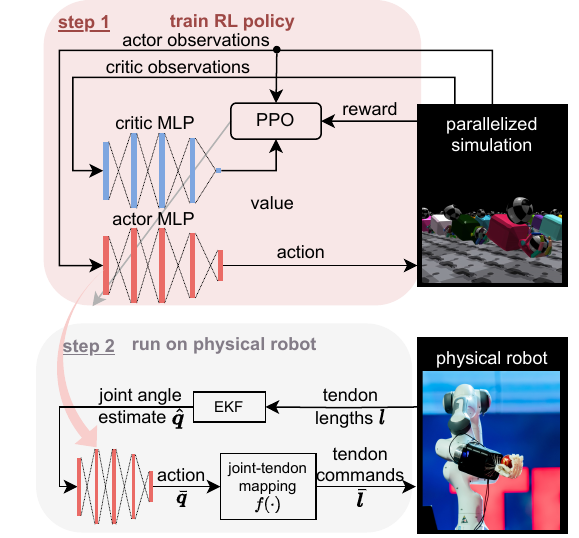}
        \caption{Overview of the RL training framework for achieving dexterous manipulation on the tendon-driven robot hand with rolling contact joints. After training the policy within a simulation environment, the actor network is transferred to the real robot.}
        \label{fig:faive_rl_sim2real}
\end{figure}

The overview of the pipeline for training the policy and running it on the real robot is shown in \cref{fig:faive_rl_sim2real}. 
The policy is trained with RL with advantage actor-critic (A2C) using asymmetric observations (where different sets of observations are given to the actor and critic). We use the PPO algorithm~\cite{Schulman2017-qc} with the implementation from the open-source repository \textit{rl\_games}~\cite{rl-games2021}. We use MLP networks as the actor and critic. 

After the training is finished, the MLP of the actor was exported to an ONNX format, a cross-compatible ML model format, and run on the robot. The joint-tendon mapping and the EKF introduced in \cref{sec:low_level_controller} were used to enable joint-level sensing and control of the Faive Hand.

\subsection{Rewards}
\Cref{table:rewards} lists the rewards used for the task and their formula.
The reward specific to the sphere rotation task is the object rotation reward, which returns the maximum reward when the rotational velocity in the $y$ axis is between $\mp 3$ and \SI{\mp 1}{\radian\per\second}, and linearly decreases outside of this region. The sign of $\omega_y$ in the reward formula can be flipped to reverse the desired direction of rotation. 

We noticed that the object's angular velocity measurement from IsaacGym contained considerable noise, which was hypothesized to be due to the collision calculation applying impulse forces whenever a part of the hand contacts the object. Computing the object rotation reward from this angular velocity measurement tended to produce policies that exploit the physics of the simulator, resulting in motions that do not actually rotate the sphere, even within the simulator. Thus we have used the numerical angular velocity for the object rotation reward, which was computed by numerically differentiating the change in object orientation between time steps.

\begin{table*}[h!]
\centering
\caption{Rewards and penalties used during training.}
\small

\begin{tabular}{c  c  c  c}
    \toprule
    \textbf{reward}         & \textbf{formula} & \textbf{weight} & \textbf{justification}  \\
    \midrule
    Object rotation           & $\min(\mp\omega_y+1, 2,\pm\omega_y + 4)$ &  0.01 & reward the rotation in the $y$ axis (can be reversed)
    \\
    Torque penalty & $||\bm \tau||_2$ & -0.02 & prevent joints from applying large torques
    \\
    Action penalty & $||\bm a ||_2$ & -0.002 & prevent large actions
    \\
    Drop penalty & $||\bm x_{obj} - \bm x_{hand}||_2 > \SI{24}{\cm}$ & -1.0 & penalize object drops  (one-time penalty after which environment is reset)
    \\
    \bottomrule
\end{tabular}

\label{table:rewards}
\end{table*}

\subsection{Observation space}
Since the actor and critic are implemented as two separate MLPs, they can be given different sets of observations. As the critic's MLP is only needed during training, it can use privileged information as input, which are data that can be obtained within the simulation but not from the real robot.
\Cref{table:observations} lists the observations used for the actor and critic. The joint positions were normalized to range between $-1$ and $1$ using the robot' joint range.
Since the measurements from the real robot contain some noise, especially for velocity measurements, we have opted to use the past five steps of the joint position measurements, in which the joint velocity is implicitly expressed.

Our target task is to rotate a sphere, which is symmetric in all axes. Due to this symmetry, the orientation of the object does not affect how the hand interacts with the object. Further, we could also remove the position measurement of the object from the actor observations without adversely affecting performance. Thus, the actor only uses proprioceptive joint data, simplifying the sim2real process and the technical challenge of obtaining accurate and low-latency measurements of the object pose. 

\begin{table}[h!]
\centering
\small
\caption{Observations used by the actor and critic networks.}
\begin{tabular}{c  c  c  c}
    \toprule
    \textbf{input}         & \textbf{dimensions} & \textbf{actor} & \textbf{critic}  \\
    \midrule
    joint pos           & 11 & & \checkmark   \\
    joint pos command   & 11 & \checkmark & \checkmark   \\
    joint pos history   & 55 & \checkmark & \\
    joint velocity      & 11 & & \checkmark   \\
    joint torque        & 11 & & \checkmark   \\
    object pos          &  3 & & \checkmark   \\
    object quat         &  4 & & \checkmark   \\
    object linear vel   &  3 & & \checkmark   \\
    object angular vel  &  3 & & \checkmark   \\
    fingertip position  & 15 & & \checkmark   \\
    fingertip quaternion& 20 & & \checkmark   \\
    fingertip lin vel   & 15 & & \checkmark   \\
    fingertip ang vel   & 15 & & \checkmark   \\
    fingertip force     & 15 & & \checkmark   \\
    previous actions    & 11 & \checkmark & \checkmark \\
    \bottomrule
\end{tabular}

\label{table:observations}
\end{table}

\subsection{Action space}
The action $\bm a$ expresses the relative change in the joint angle command. It is first clipped to between $-1$ and $1$, then it increments the desired joint angle $\bar{\bm q}$ as
\begin{equation}
    \bar{\bm q} \leftarrow \text{clip}(\bar{\bm q} + v_{max} \Delta t \bm a , \bm q_{min}, \bm q_{max})
\end{equation}
where $\Delta t$ is the timestep and $v_{max}$ is a constant scalar that caps the maximum speed of the joints from the policy. $v_{max}$ was set to \SI{5}{\radian\per\second}. After the action updates the desired joint angle, it is clipped between the minimum $\bm q_{min}$ and maximum $\bm q_{max}$ joint angles of the robot hand.

\subsection{Domain randomization}
To compensate for the inaccuracy of the physics engine and to make the policy more robust to overcome the sim2real gap, domain randomization was applied to the physics properties, namely the observations, damping and stiffness of the tendons and the joints, joint range of motion, mass and friction of the robot and object, and object scale.

\section{Experiments}
\subsection{Training the policy in simulation}
The GPU-based high-performance physics simulator IsaacGym \cite{Makoviychuk2021-ll} was used to simulate the robot for training the policy. 4096 environments were simulated in parallel on a single NVIDIA A10G GPU. The simulation was run at 60 Hz, while the policy ran every three steps, resulting in a 20 Hz policy. The actor and critic networks were separate MLPs with 4 hidden layers of dimensions $ [512, 512, 256, 128] $  and ELU activations. We have trained with and without domain randomization (DR) to compare its effect on the performance of the policy on the real robot.

\begin{figure}
        \centering
        \includegraphics[width=\columnwidth/50*24]{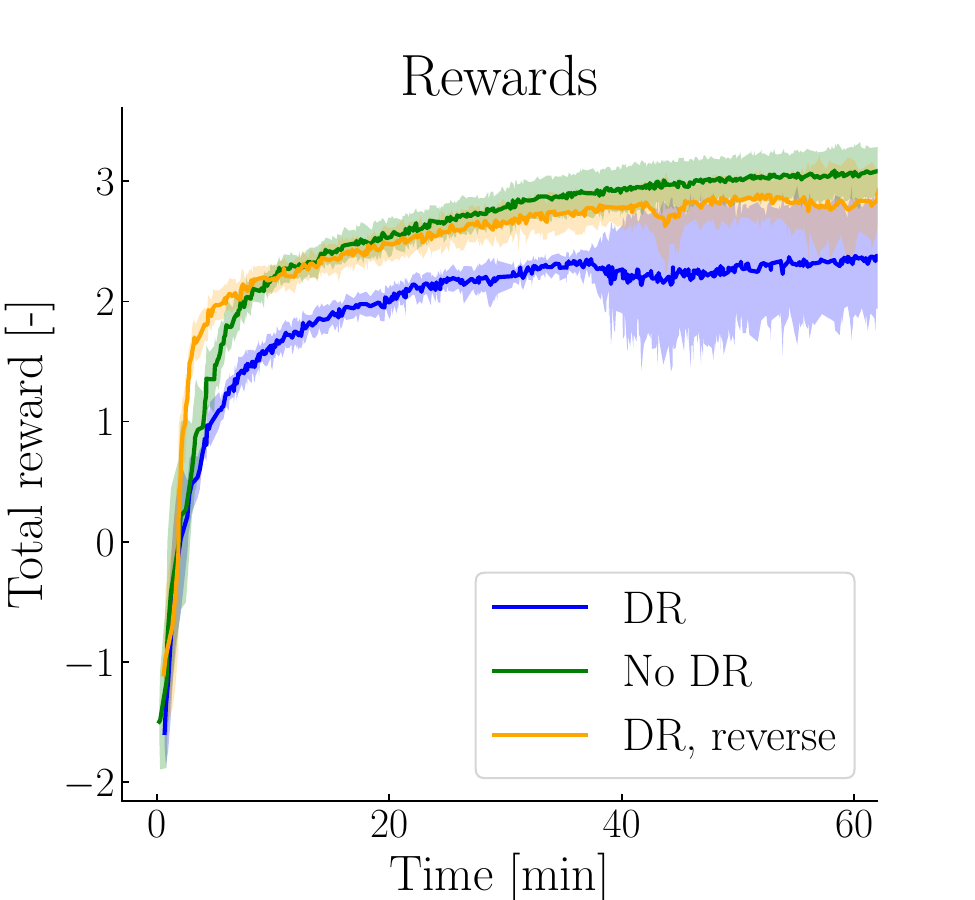}
        \includegraphics[width=\columnwidth/50*24]{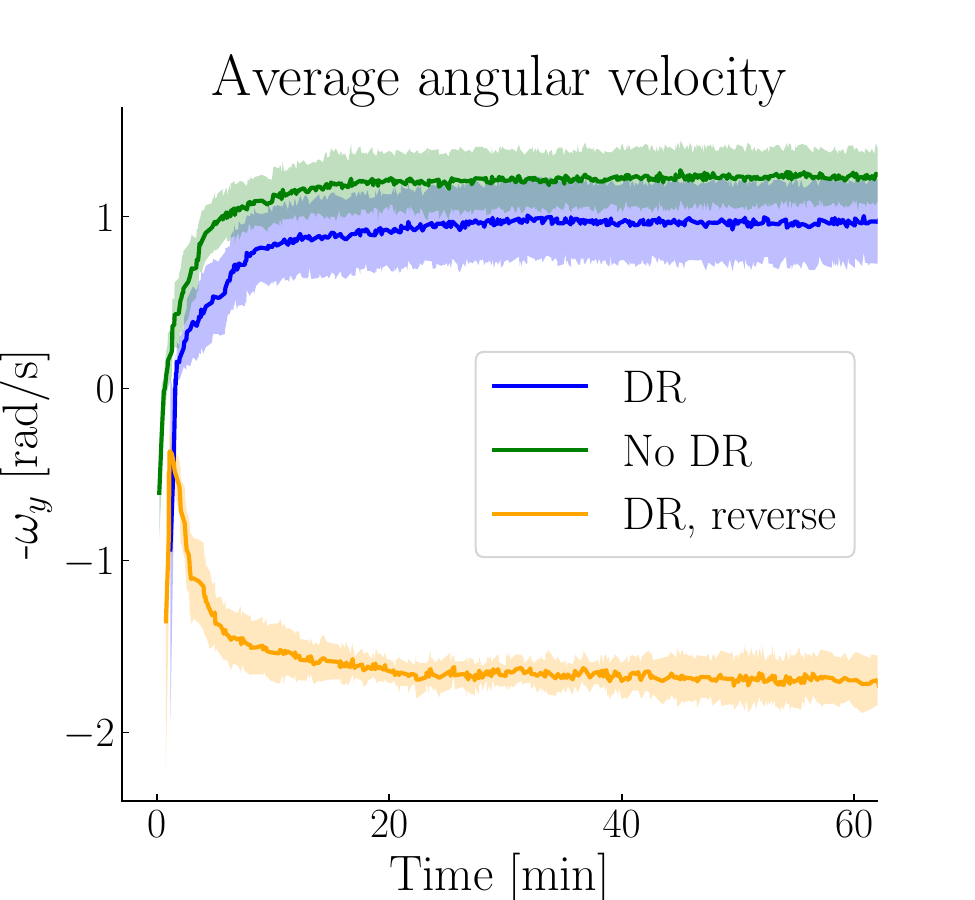}
        
        \caption{Training curve evolution for the policies, trained with and without domain randomization (DR), and for a reversed target rotation direction. We took the mean of 9 training rounds for each approach. An area of $\pm\sigma$ is shown around both plots.}
        \label{fig:rl_training_curve}
\end{figure}

\Cref{fig:rl_training_curve} shows the resulting training curve. As RL performance heavily depends on random seed selection, we have run the training with multiple random seeds and plotted their mean and standard deviation. As an additional indicator of the policy performance besides the reward, we have also logged the angular velocity in the desired direction along the $y$ axis.

We have also tried to rotate the ball in the $x$ and $z$ axes as well, by swapping the $\omega_y$ in the reward to $\omega_x$ and $\omega_z$. However the RL algorithm was not able to come up with a policy to consistently rotate the ball in those axes. We attribute this phenomenon to the lack of joints for abduction and adduction on this hand prototype, which are the joints to spread fingers apart and together. When we ourselves try to rotate objects in the $x$ and $y$ axes in our hands as well, we see that these joints are required. 

It can be seen in \cref{fig:rl_training_curve} that the performance rapidly increases in the first 10 minutes then gradually plateaus out afterwards. For one of the training runs with DR enabled, the policy performance suddenly collapsed at around 30 minutes, which increases the standard deviation of the reward and slightly lowers the mean. This suggests that testing multiple runs with different random seeds is crucial for evaluating the actual performance of the RL setup. For both metrics logged, the performance is higher when DR is disabled. This is to be expected as the non-DR policy can be fine-tuned to an environment with a single set of physics parameters, where as the DR policy must work with many different physics parameters.

\begin{figure*}
        \centering
        \includegraphics[trim={0mm 0mm 0mm 12mm}, clip, width=1.8\columnwidth]{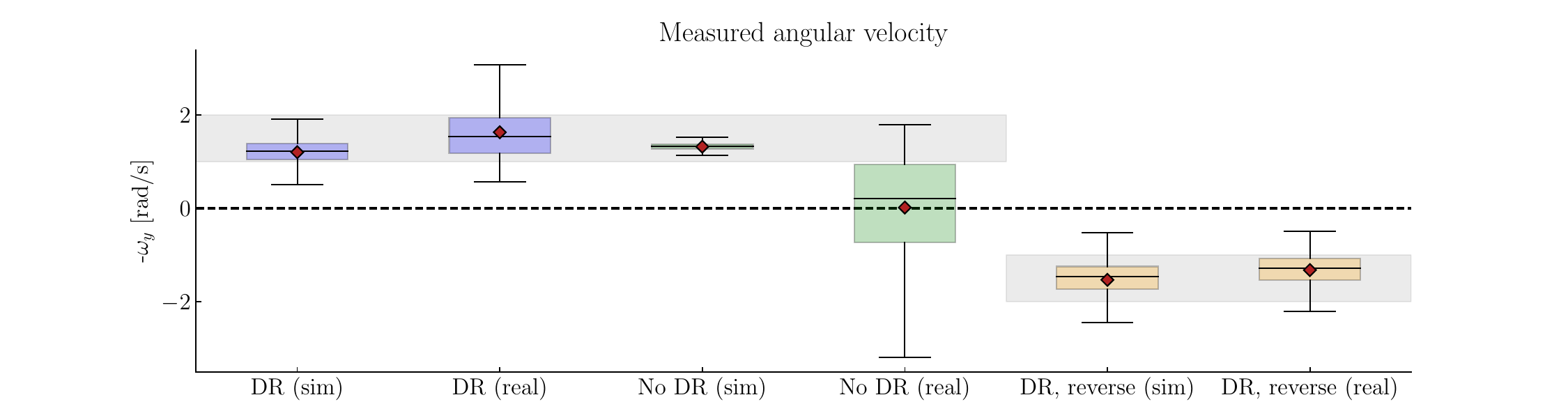}
        \caption{Distribution of the object rotational velocity on the real and simulated robot, for policies trained with and without DR. The gray strip indicates the region in which the object rotation reward is at its maximum value. The diamond indicates the mean.}
        \label{fig:result_real_robot}
\end{figure*}

\subsection{Running the policy on the real robot}
The performance difference across different random seeds was even larger when it was applied to the real robot than it was within the simulation. Some policies even stopped moving the finger after a few seconds of running the policy, getting stuck in a hand pose where the policy outputs zero values for actions.
Therefore, we ran each of the policies trained with separate random seeds, and picked the best performing one from each condition to evaluate the best-case performance for each.
Interestingly, we have found that multiplying the joint position measurements by $0.5$ before sending it to the policy improves the performance of the policy when run on the real robot, and achieves a motion much closer to that in simulation. We hypothesize that as the finger presses down on the ball, the tendon stretches and the structure deforms, pulling the tendon further than it was at the moment of contact. This affects the joint angle estimate from the EKF, which erroneously estimates that the finger is bent more than it actually is. We believe that scaling down the observations works as a simple corrective measure for this effect.

To measure the rotation of the ball, we have embedded an IMU and Bluetooth device (Arduino Nano 33 BLE) within a spherical \textit{gachapon} capsule toy ball, and received the orientation (obtained by fusing accelerometer and gyroscopic measurements with the Madgwick filter) and rotational velocity of the object via Bluetooth. These measurements were not used in the policy, and were used just for evaluating the performance. The rotational velocity was converted to the robot's frame, and smoothed with an exponential smoothing filter to remove noise from measurements.

The result is shown in \cref{fig:result_real_robot}. The gray strip indicates the region in which the object rotation reward is at its maximum value, \textit{i.e.} the "target region" of the policy.
Within simulation, the mean of the angular velocity of the object is similar in the DR and non-DR policies. However, the DR policy exhibits a larger distribution in the velocity, presumably due to the variance in performance due to the randomized physics parameters.
When they are applied to the real robot, the importance of DR becomes apparent, as the non-DR policy fails to rotate the sphere, just rocking it back and forth in its hand. The DR policy succeeds in consistently rotating the ball, achieving the target rotational velocity for the majority of the measurements.
The policy trained with a reversed rotational direction was also applied to the real robot, which again successfully rotated the ball to within the target velocity, for the majority of the sampled measurements. \Cref{fig:humanoids_overview} shows three snapshots, taken 2 seconds apart, of the policy running on the robot, and the accompanying video also shows videos of each policy running on the real robot. 

\section{Conclusion}
We have introduced our anthropomorphic hand platform for use in autonomous manipulation.
For this platform, we have developed a method to model, control, and sense rolling contact joints so that they can be integrated into a parallelized simulation environment to train a closed-loop policy. We show that the trained policy can be run on our physical robotic hand to achieve dexterous sphere rotation.

While simulators and environments for RL training for dexterous manipulation have become more capable and accessible recently, we have yet to see the same for five-fingered biomimetic robotic hands. The Faive Hand developed at the Soft Robotics Lab is aimed at making dexterous manipulators more capable and accessible. In this work we show that the hand can achieve zero-shot transfer of skills trained with RL in the IsaacGym simulator, showing the potential of this hand to be used for other tasks trained with RL.

However, there are still limitations in the software and hardware. When we have tried to apply cube reorientation tasks such as in \cite{OpenAI2018-te}, the policy did work in simulation, but failed on the real robot. This motion is more complex than our single-axis sphere rotation task, as the object is not symmetric as the sphere and must be rotated around all three axes. We attribute the failure on the real robot to a multitude of factors, such as poor joint angle measurement from the EKFs, especially when there is contact, and a lack of proper system identification to ensure accurate actuation dynamics in the IsaacGym simulator. We will continue to develop the robotic hand to solve these challenges, with a combination of physical and programmatic approaches, such as integrating more sensors for proprioceptive measurements or a better system identification process to reduce the sim2real gap. 

\addtolength{\textheight}{-8cm}   %

\section*{ACKNOWLEDGMENT}
The authors thank Jonas Lauener for creating the two-finger prototype of the rolling contact joint finger.
Yasunori Toshimitsu is partially funded by the Takenaka Scholarship Foundation, the Max Planck ETH Center for Learning Systems, and the Swiss Government Excellence Scholarship.
This work was partially funded by the Amazon Research Awards.  This work was also supported by an ETH RobotX research grant funded through the ETH Zurich Foundation.

\bibliography{IEEEabrv,IEEEexample,paperpile,websites}

\begin{thebibliography}{10}
\providecommand{\url}[1]{#1}
\csname url@rmstyle\endcsname
\providecommand{\newblock}{\relax}
\providecommand{\bibinfo}[2]{#2}
\providecommand\BIBentrySTDinterwordspacing{\spaceskip=0pt\relax}
\providecommand\BIBentryALTinterwordstretchfactor{4}
\providecommand\BIBentryALTinterwordspacing{\spaceskip=\fontdimen2\font plus
\BIBentryALTinterwordstretchfactor\fontdimen3\font minus
  \fontdimen4\font\relax}
\providecommand\BIBforeignlanguage[2]{{%
\expandafter\ifx\csname l@#1\endcsname\relax
\typeout{** WARNING: IEEEtran.bst: No hyphenation pattern has been}%
\typeout{** loaded for the language `#1'. Using the pattern for}%
\typeout{** the default language instead.}%
\else
\language=\csname l@#1\endcsname
\fi
#2}}

\bibitem{OpenAI2018-te}
{OpenAI}, M.~Andrychowicz, B.~Baker, M.~Chociej, R.~Jozefowicz, B.~McGrew,
  J.~Pachocki, A.~Petron, M.~Plappert, G.~Powell, A.~Ray, J.~Schneider,
  S.~Sidor, J.~Tobin, P.~Welinder, L.~Weng, and W.~Zaremba, ``Learning
  dexterous in-hand manipulation,'' Aug. 2018.

\bibitem{Makoviychuk2021-ll}
V.~Makoviychuk, L.~Wawrzyniak, Y.~Guo, M.~Lu, K.~Storey, M.~Macklin,
  D.~Hoeller, N.~Rudin, A.~Allshire, A.~Handa, and {Gavriel State}, ``Isaac
  gym: High performance {GPU} based physics simulation for robot learning,''
  Nov. 2021.

\bibitem{Rudin2022-xe}
N.~Rudin, D.~Hoeller, P.~Reist, and M.~Hutter, ``Learning to walk in minutes
  using massively parallel deep reinforcement learning,'' in \emph{Proceedings
  of the 5th Conference on Robot Learning}, ser. Proceedings of Machine
  Learning Research, A.~Faust, D.~Hsu, and G.~Neumann, Eds., vol. 164.\hskip
  1em plus 0.5em minus 0.4em\relax PMLR, 2022, pp. 91--100.

\bibitem{Handa2022-sn}
A.~Handa, A.~Allshire, V.~Makoviychuk, A.~Petrenko, R.~Singh, J.~Liu,
  D.~Makoviichuk, K.~Van~Wyk, A.~Zhurkevich, B.~Sundaralingam, Y.~Narang, J.-F.
  Lafleche, D.~Fox, and {Gavriel State}, ``{DeXtreme}: Transfer of agile
  in-hand manipulation from simulation to reality,'' Oct. 2022.

\bibitem{Chen2022-ag}
T.~Chen, J.~Xu, and P.~Agrawal, ``A system for general in-hand object
  re-orientation,'' in \emph{Proceedings of the 5th Conference on Robot
  Learning}, ser. Proceedings of Machine Learning Research, A.~Faust, D.~Hsu,
  and G.~Neumann, Eds., vol. 164.\hskip 1em plus 0.5em minus 0.4em\relax PMLR,
  2022, pp. 297--307.

\bibitem{Yin2023-kv}
Z.-H. Yin, B.~Huang, Y.~Qin, Q.~Chen, and X.~Wang, ``Rotating without seeing:
  Towards in-hand dexterity through touch,'' Mar. 2023.

\bibitem{Allshire2022-ik}
A.~Allshire, M.~MittaI, V.~Lodaya, V.~Makoviychuk, D.~Makoviichuk, F.~Widmaier,
  M.~Wüthrich, S.~Bauer, A.~Handa, and A.~Garg, ``Transferring dexterous
  manipulation from {GPU} simulation to a remote real-world {TriFinger},'' in
  \emph{2022 IEEE/RSJ International Conference on Intelligent Robots and
  Systems (IROS)}, Oct. 2022, pp. 11\,802--11\,809.

\bibitem{WonikRobotics2023}
\BIBentryALTinterwordspacing
W.~Robotics, ``Allegro hand: Highly adaptive robotic hand for r\&d,'' 2023.
  [Online]. Available: \url{https://www.wonikrobotics.com/research-robot-hand}
\BIBentrySTDinterwordspacing

\bibitem{ShadowRobot2023}
\BIBentryALTinterwordspacing
S.~R. Company, ``Shadow dexterous hand series - research and development
  tool,'' 2023. [Online]. Available:
  \url{https://www.shadowrobot.com/dexterous-hand-series/}
\BIBentrySTDinterwordspacing

\bibitem{Shaw_undated-sg}
K.~Shaw, A.~Agarwal, and D.~Pathak, ``{LEAP} hand: Low-cost, efficient, and
  anthropomorphic hand for robot learning,''
  \url{https://www.roboticsproceedings.org/rss19/p089.pdf}, accessed:
  2023-7-11.

\bibitem{Gurtler2023-hv}
N.~Gürtler, F.~Widmaier, C.~Sancaktar, S.~Blaes, P.~Kolev, S.~Bauer,
  M.~Wüthrich, M.~Wulfmeier, M.~Riedmiller, A.~Allshire, Q.~Wang, R.~McCarthy,
  H.~Kim, J.~B. Pohang, W.~Kwon, S.~Qian, Y.~Toshimitsu, M.~Y. Michelis,
  A.~Kazemipour, A.~Raayatsanati, H.~Zheng, B.~G. Cangan, B.~Schölkopf, and
  G.~Martius, ``Real robot challenge {2022}: Learning dexterous manipulation
  from offline data in the real world,'' Aug. 2023.

\bibitem{Wuthrich2021-cb}
M.~Wuthrich, F.~Widmaier, F.~Grimminger, S.~Joshi, V.~Agrawal, B.~Hammoud,
  M.~Khadiv, M.~Bogdanovic, V.~Berenz, J.~Viereck, M.~Naveau, L.~Righetti,
  B.~Schölkopf, and S.~Bauer, ``{TriFinger}: An open-source robot for learning
  dexterity,'' in \emph{Proceedings of the 2020 Conference on Robot Learning},
  ser. Proceedings of Machine Learning Research, J.~Kober, F.~Ramos, and
  C.~Tomlin, Eds., vol. 155.\hskip 1em plus 0.5em minus 0.4em\relax PMLR, 2021,
  pp. 1871--1882.

\bibitem{Shi2021-gy}
F.~Shi, T.~Homberger, J.~Lee, T.~Miki, M.~Zhao, F.~Farshidian, K.~Okada,
  M.~Inaba, and M.~Hutter, ``Circus {ANYmal}: A quadruped learning dexterous
  manipulation with its limbs,'' in \emph{2021 IEEE International Conference on
  Robotics and Automation (ICRA)}.\hskip 1em plus 0.5em minus 0.4em\relax IEEE,
  May 2021.

\bibitem{Salisbury1982-xs}
J.~K. Salisbury and J.~J. Craig, ``Articulated hands: Force control and
  kinematic issues,'' \emph{Int. J. Rob. Res.}, vol.~1, no.~1, pp. 4--17, Mar.
  1982.

\bibitem{Kawaharazuka2020-ff}
K.~Kawaharazuka, S.~Makino, M.~Kawamura, S.~Nakashima, Y.~Asano, K.~Okada, and
  M.~Inaba, ``Human mimetic forearm and hand design with a radioulnar joint and
  flexible machined spring finger for human skillful motions,'' \emph{Journal
  of Robotics and Mechatronics}, vol.~32, no.~2, pp. 445--458, 2020.

\bibitem{Makino2018-ki}
S.~Makino, K.~Kawaharazuka, A.~Fujii, M.~Kawamura, T.~Makabe, M.~Onitsuka,
  Y.~Asano, K.~Okada, K.~Kawasaki, and M.~Inaba, ``Five-fingered hand with wide
  range of thumb using combination of machined springs and variable stiffness
  joints,'' in \emph{2018 IEEE/RSJ International Conference on Intelligent
  Robots and Systems (IROS)}, Oct. 2018, pp. 4562--4567.

\bibitem{Puhlmann2022-yi}
S.~Puhlmann, J.~Harris, and O.~Brock, ``{RBO} hand 3: A platform for soft
  dexterous manipulation,'' \emph{IEEE Trans. Rob.}, vol.~38, no.~6, pp.
  3434--3449, Dec. 2022.

\bibitem{Schlagenhauf2018-xt}
C.~Schlagenhauf, D.~Bauer, K.-H. Chang, J.~P. King, D.~Moro, S.~Coros, and
  N.~Pollard, ``Control of tendon-driven soft foam robot hands,'' in \emph{2018
  IEEE-RAS 18th International Conference on Humanoid Robots (Humanoids)}, Nov.
  2018, pp. 1--7.

\bibitem{Kim2023-cx}
S.~Kim, E.~Sung, and J.~Park, ``{ARC} joint: Anthropomorphic rolling contact
  joint with kinematically variable torsional stiffness,'' \emph{IEEE Robotics
  and Automation Letters}, vol.~8, no.~3, pp. 1810--1817, Mar. 2023.

\bibitem{Kim2019-vc}
Y.-J. Kim, J.~Yoon, and Y.-W. Sim, ``Fluid lubricated dexterous finger
  mechanism for human-like impact absorbing capability,'' \emph{IEEE Robot.
  Autom. Lett.}, vol.~4, no.~4, pp. 3971--3978, Oct. 2019.

\bibitem{noauthor_undated-ir}
S.~W. Hong, J.~Yoon, Y.-J. Kim, and H.~S. Gong, ``\BIBforeignlanguage{en}{Novel
  implant design of the proximal interphalangeal joint using an optimized
  rolling contact joint mechanism},'' \emph{\BIBforeignlanguage{en}{J. Orthop.
  Surg. Res.}}, vol.~14, no.~1, p. 212, July 2019.

\bibitem{Kim2016-fl}
S.-H. Kim, H.~In, J.-R. Song, and K.-J. Cho, ``Force characteristics of rolling
  contact joint for compact structure,'' in \emph{2016 6th IEEE International
  Conference on Biomedical Robotics and Biomechatronics (BioRob)}, June 2016,
  pp. 1207--1212.

\bibitem{Ookubo2015-bc}
S.~Ookubo, Y.~Asano, T.~Kozuki, T.~Shirai, K.~Okada, and M.~Inaba, ``Learning
  nonlinear muscle-joint state mapping toward geometric model-free tendon
  driven musculoskeletal robots,'' in \emph{2015 IEEE-RAS 15th International
  Conference on Humanoid Robots (Humanoids)}, Nov. 2015, pp. 765--770.

\bibitem{10.7717/peerj-cs.103}
\BIBentryALTinterwordspacing
A.~Meurer, C.~P. Smith, M.~Paprocki, O.~\v{C}ert\'{i}k, S.~B. Kirpichev,
  M.~Rocklin, A.~Kumar, S.~Ivanov, J.~K. Moore, S.~Singh, T.~Rathnayake,
  S.~Vig, B.~E. Granger, R.~P. Muller, F.~Bonazzi, H.~Gupta, S.~Vats,
  F.~Johansson, F.~Pedregosa, M.~J. Curry, A.~R. Terrel, v.~Rou\v{c}ka,
  A.~Saboo, I.~Fernando, S.~Kulal, R.~Cimrman, and A.~Scopatz, ``Sympy:
  symbolic computing in python,'' \emph{PeerJ Computer Science}, vol.~3, p.
  e103, Jan. 2017. [Online]. Available:
  \url{https://doi.org/10.7717/peerj-cs.103}
\BIBentrySTDinterwordspacing

\bibitem{Schulman2017-qc}
J.~Schulman, F.~Wolski, P.~Dhariwal, A.~Radford, and O.~Klimov, ``Proximal
  policy optimization algorithms,'' July 2017.

\bibitem{rl-games2021}
D.~Makoviichuk and V.~Makoviychuk, ``rl-games: A high-performance framework for
  reinforcement learning,'' \url{https://github.com/Denys88/rl_games}, May
  2021.

\end{thebibliography}
\bibliographystyle{IEEEtran}

\end{document}